# An Amharic News Text classification Dataset

**Israel Abebe Azime**
AIMS-AMMI
iazime@aimsammi.com

**Nebil Mohammed**
Addis Ababa Science and Technology University
nebilmohammedsaw12@gmail.com

## Abstract

In NLP, text classification is one of the primary problems we try to solve and its uses in language analyses are indisputable. The lack of labeled training data made it harder to do these tasks in low resource languages like Amharic. The task of collecting, labeling, annotating, and making valuable this kind of data will encourage junior researchers, schools, and machine learning practitioners to implement existing classification models in their language.

In this short paper, we aim to introduce the Amharic text classification dataset that consists of more than 50k news articles that were categorized into 6 classes. This dataset is made available with easy baseline performances to encourage studies and better performance experiments.

[GitHub Link](#)

## 1 Amharic Language

Amharic is the second most spoken Semitic language. It is the official working language of 100 million people that reside in the Federal Democratic Republic of Ethiopia. The language uses its unique alphabet called Fidel. Amharic alphabet consists of punctuation and numbers in addition to its 231 primary letters(Gezmu et al., 2018).

Amharic is considered as a low resource language (Gezmu et al., 2018). This is not due to the lack of raw data, rather it is due to the scarcity of labeled data. Most of the time researchers prepare data for their use but fail to make the dataset available.

## 2 Introduction

Text classification or text categorization is a task of assigning a sentence, paragraphs or documents into one of $n$ classes we have on our dataset. This task is one of the core NLP tasks that needs manually annotated data as an input(Kowsari et al., 2019).

Tasks like Sentiment analysis, News categorization, Topic Analysis and more are prominent application of classification task (Kowsari et al., 2019). we usually use languages like English for NLP tasks especially in academia for education and we don't study the effect of different algorithms in languages which have different structure than English. This does not consider characteristics of low resource languages while developing new algorithms.

## 3 Previous works

Text classification task is one of the core NLP tasks that needs manually annotated data as an input(Kowsari et al., 2019).

There are some works done by (Sahle-mariam and Daniel, 2009), (TEGEGNIE, 2010), (Kelemework, 2013), (Seffi Gebeyehu, 2014) and others. We have found that all of them have used a very small dataset which ranges from 200 - 15,000 articles from a single data source. Some researches also talk about lack of standard Amharic text classification corpus (TEGEGNIE, 2010).

## 4 Text Classification Dataset

Our dataset consists of 6 classes. This class information was found by the tag we get from websites and we manually verify the case and removed noises in the process.

Table 3 shows the detailed description of the web pages we collected our dataset from. The web pages are local and international news sites. We have collected the datasets from different sources to increase the variety of the text. As far as we know, this is the first work with 1) data collected from different sources, 2) data size is at least 5 times greater than the existing benchmark datasets.

We included several details that might be useful for different purposes. It includes information like the web page the article is found from, Views it had, the title of the article, and the date it was posted at. However, the 'category' and 'article' metadata are very important while the other metadata might be still useful for different use-cases.

In this data local news and international news refers to topics that are not included in the rest and are categorized as international or regional issue.

| Class Name | items |
|---|---|
| ሀገር አቀፍ ዜና / local news | 20564 |
| ስፖርት / sport | 9812 |
| ፖለቲካ / politics | 9307 |
| ዓለም አቀፍ ዜና / international news | 6515 |
| ቢዝነስ / business | 3873 |
| መዝናኛ / entertainment | 635 |
| Total | 50706 |

Table 1: Different class distrbution in the dataset

## 5 Baseline Performance

We performed simple classification training tasks on our dataset. We used simpler models like Naive Bayes classifier using Count vectorizer and TF-IDF features as can be seen in Table 2.

Our dataset pre-processing includes simple white space tokenization with removing punctuation marks and applying character level normalization.

Character level normalization in this context refers to replacing characters that have the same sound and can be used interchangeably. This will correct the occurrence of similar words multiple times in our vocabulary and be considered as different once.

## 6 CONCLUSION

In this work, we release the Amharic text classification dataset. This work contributes to the low amount of Amharic text classification dataset and aims to be a good starting point for future Amharic text classification works.

| Baseline Model | Accuracy in % |
|---|---|
| Naive Bayes using count vectorizer features | 62.2 % |
| Naive Bayes using Tf-idf features | 62.3 % |

Table 2: Baseline classification performance on the data.

To help the Amharic ML community we have put the data and baseline scores in the GitHub repository we provided.

Future works in this dataset include trying to improve the performance of the models using advanced word embedding and transformer models. From the table 1, you can see our dataset is imbalanced so we are thinking of using text augmentation techniques to increase dataset size. Different approaches to imbalance class problems should be explored on this dataset too. There are a lot of approaches that we know and use for data imbalance and exploring those approaches on this dataset can show us the effectiveness of the methods in Amharic language data.

| News Site | Items | URL |
|---|---|---|
| Addis Admas | 1839 | www.addisadmassnews.com |
| Addis Maleda | 847 | www.addismaleda.com |
| Al-Ain Amharic | 887 | am.al-ain.com/ |
| Amhara MM | 2438 | www.amharaweb.com/ |
| BBC Amharic | 816 | www.bbc.com/amharic/ |
| Ethiopian Press | 5597 | www.press.et/Ama/ |
| Ethiopian Reporter | 6280 | www.ethiopianreporter.com |
| Fana Broadcasting | 7700 | www.fanabc.com |
| Soccer Ethiopia | 8595 | soccerethiopia.net |
| VOA Amharic | 6943 | amharic.voanews.com |
| Walta | 8764 | waltainfo.com/am/ |
| Total | 50706 | |

Table 3: Dataset Source distrbution

## References

Andargachew Mekonnen Gezmu, B. Seyoum, M. Gasser, and A. Nürnberger. 2018. Contemporary amharic corpus: Automatically morpho-syntactically tagged amharic corpus.

Worku Kelemework. 2013. Automatic amharic text news classification: Aneural networks approach. *Ethiop. J. Sci. & Technol. 6(2) 127-137, 2013.*

Kamran Kowsari, Kiana Jafari Meimandi, Mojtaba Heidarysafa, Sanjana Mendu, Laura E. Barnes, and Donald E. Brown. 2019. Text classification algorithms: A survey. *CoRR preprint arXiv:1904.08067*.

Mulugeta Sahlemariam, Meron; Libsie and Yacob Daniel. 2009. Concept-based automatic amharic document categorization. *AMCIS 2009 Proceedings. 116. https://aisel.aisnet.org/amcis2009/116*.

Dr.Vuda Sreenivasa Rao2 Seffi Gebeyehu. 2014. A two step data mining approach for amharic text classification. *American Journal of Engineering Research (AJER) e-ISSN : 2320-0847 p-ISSN : 2320-0936*.

ALEMU KUMILACHEW TEGEGNIE. 2010. Hierarchical amharic news text classification hierarchical amharic news text classification.